%% file: iclr2026_conference.tex
\documentclass{article} 
\usepackage{iclr2026_conference,times}

\input{math_commands.tex}

\input{macros}

\input{packages}

\title{Semi-structured LLM Reasoners Can Be Rigorously Audited}

\author{Jixuan Leng \quad Cassandra A. Cohen \quad Zhixian Zhang \\
\textbf{Chenyan Xiong} \quad \textbf{William W. Cohen}
\\
Carnegie Mellon University\ \ \ 
\\
\texttt{jixuanl, wcohen@cs.cmu.edu}
}

\iclrfinalcopy 
\begin{document}

\maketitle

\begin{abstract}
Although Large Language Models (LLMs) have become  capable  reasoners, the problem of faithfulness persists: their reasoning can contain errors and omissions that are difficult to detect and that may obscure biases in model outputs.
To address this issue, we introduce Semi-Structured Reasoning Models (SSRMs), which are trained to produce semi-structured representations of reasoning.
SSRMs generate reasoning traces in a \emph{non-executable} Pythonic syntax that names each reasoning step and marks its inputs and outputs. 
This structure allows SSRM traces to be automatically \emph{audited} to identify reasoning flaws.
We evaluate three types of audits: hand-crafted \emph{structured reasoning audits}, written in a domain-specific language (DSL) implemented in Python; LLM-generated \emph{structured reasoning audits}; and learned \emph{typicality audits}, which apply probabilistic models over reasoning traces. 
We show that all of these methods can be used to effectively flag probable reasoning errors.
Importantly, the auditability of SSRMs does not appear to compromise overall accuracy: in evaluation on twelve benchmarks and two model families, SSRMs demonstrate strong performance and generalizability relative to other models of comparable size.
We have provided our code publicly at \href{https://github.com/wwcohen/doctest-prompting}{Github Link}.
\end{abstract}

\input{sections/introduction}
\input{sections/related_work}

\input{sections/methods}
\input{sections/experiment}

\input{sections/conclusion}

\clearpage

\ificlrfinal
\else
\section*{Reproducibility Statement} 
To facilitate reproducibility, we provide detailed information on the datasets used (please see Appendix~\ref{appendix_evaluation_datasets}), implementation details (please see Appendix~\ref{appendix_training_details}), and code (\href{https://anonymous.4open.science/r/ICLR2026-DCAC}{Anonymous Github Link}).
\fi

{
\small
\bibliographystyle{iclr2026_conference}
\bibliography{iclr2026_conference}
}

\clearpage
\appendix
\input{sections/appendix}

\end{document}

%% file: math_commands.tex
\usepackage{amsmath,amsfonts,bm}

\def\eqref#1{equation~\ref{#1}}

\def\1{\bm{1}}

\DeclareMathAlphabet{\mathsfit}{\encodingdefault}{\sfdefault}{m}{sl}
\SetMathAlphabet{\mathsfit}{bold}{\encodingdefault}{\sfdefault}{bx}{n}



%% file: macros.tex
\newcommand{\ignorethis}[1]{}

\makeatletter
\DeclareRobustCommand\onedot{\futurelet\@let@token\@onedot}
\def\@onedot{\ifx\@let@token.\else.\null\fi\xspace}

\makeatother

\makeatletter
\def\adl@drawiv#1#2#3{%
        \hskip.5\tabcolsep
        \xleaders#3{#2.5\@tempdimb #1{1}#2.5\@tempdimb}%
                #2\z@ plus1fil minus1fil\relax
        \hskip.5\tabcolsep}
\newcommand{\cdashlinelr}[1]{%
  \noalign{\vskip\aboverulesep
           \global\let\@dashdrawstore\adl@draw
           \global\let\adl@draw\adl@drawiv}
  \cdashline{#1}
  \noalign{\global\let\adl@draw\@dashdrawstore
           \vskip\belowrulesep}}
\makeatother

\definecolor{highlight}{HTML}{71b7ed}
\definecolor{second}{HTML}{B8DBF6}
\definecolor{customBlue}{RGB}{113,183,237}
\definecolor{customGreen}{RGB}{113,237,183}

\newcommand{\bb}[1]{\textbf{#1}}
\newcommand{\ii}[1]{\textit{#1}}
\newcommand{\uu}[1]{\underline{#1}}

%% file: packages.tex
\usepackage[inkscapelatex=false]{svg}

\usepackage{color,xcolor}
\usepackage{epsfig}
\usepackage{graphicx,makecell}
\usepackage{graphics}
\usepackage{inconsolata}
\usepackage{lineno}
\usepackage{xstring}

\usepackage{hyperref}
\usepackage{url}

\usepackage{adjustbox}
\usepackage{array}
\usepackage{booktabs}
\usepackage{colortbl}
\usepackage{float,wrapfig}
\usepackage{hhline}
\usepackage{multirow}
\usepackage{subcaption} 
\usepackage[toc,page]{appendix}
\usepackage{tabularx}
\usepackage{tcolorbox}

\usepackage{bbm}
\usepackage{amsmath,amsfonts,amsthm,amssymb}
\usepackage{bm}
\usepackage{nicefrac}
\usepackage{microtype}
\usepackage{inconsolata}
\usepackage{pifont}
\usepackage{dsfont}
\usepackage{fontawesome5}

\usepackage{changepage}
\usepackage{extramarks}
\usepackage{fancyhdr}
\usepackage{lastpage}
\usepackage{setspace}
\usepackage{soul}
\usepackage{xspace}
\usepackage{indentfirst}
\usepackage{paralist}

\usepackage{algorithm}
\usepackage{enumerate}
\usepackage{lipsum}
\usepackage{physics}
\usepackage{cleveref}

\usepackage{listings}
\usepackage{xcolor}
\usepackage{enumitem}
\usepackage{fancyvrb}
\usepackage{fvextra}
\usepackage{algpseudocode}
\usepackage{placeins}

\definecolor{commentgray}{rgb}{0.5, 0.5, 0.5}

%% file: sections/introduction.tex
\section{Introduction}

Large Language Models (LLMs) often benefit from reasoning techniques such as short Chain-of-Thought (CoT) prompting~\citep{wei2022chain} or long CoT reasoning~\citep{chen2025towards,wang2025think,wang2025tutorial}.
Yet in many applications, LLMs may generate superficially plausible but incorrect reasoning that obscures biases in the output~\citep{turpin2024language}; more generally,  reasoning traces are not causally related to the final output~\citep{bao2024llms}.
This problem of  ``unfaithful''LLM  reasoning has been extensively investigated in short CoT settings~\citep{lanham2023measuring,bentham2024chainofthoughtunfaithfulnessdisguisedaccuracy,parcalabescu2024measuringfaithfulnessselfconsistencynatural}, and is likely to be more problematic in long CoT reasoning.

\input{figures/problem}

As a concrete step toward demystifying reasoning LLMs and improving their reliability, we present methods for \emph{rigorously checking LLM reasoning on specific tasks}.
To illustrate and motivate this, consider the simplified medical question-answering (QA) task in Figure~\ref{fig:problem}, adapted from the MedCalcBench~\citep{khandekar2024medcalc}.
The ``flawed'' reasoning trace appears superficially plausible but is incomplete compared to the ``ideal'' trace: it contains one obvious omission, one subtler error, and one issue where the LLM fails to explicitly check the compatibility of the units for an extracted value.
Although none of these affect the final answer in this example, such flaws are undesirable in consequential tasks.
Human experts performing similar tasks are often expected to carefully follow explicit instructions---variously called rubrics, cookbooks, or policies depending on the domain---to ensure that reasoning is complete and decisions are made consistently.  
This observation motivates the central research question: \emph{can we detect when an LLM deviates from a desired reasoning strategy?}

\input{figures/approach}

Since analyzing arbitrary reasoning traces is difficult, 
we begin by training an LLM to generate \emph{semi-structured} reasoning traces, as shown in Figure~\ref{fig:approach} (Top). 
Following prior work~\citep{cohen2024watch}, we adopt a Pythonic syntax that labels different types of reasoning steps using a restricted, task-specific vocabulary and specifies the \emph{inputs and outputs} of each step, \emph{without requiring the steps to be executable}. 
Because the steps can perform arbitrary computations and consume or produce arbitrary strings of text, this semi-structured format is highly general.
In this paper we
provide new evidence for this generality by showing that training models to generate \emph{semi-structured} traces achieves performance comparable to similarly trained free-form reasoning models and other baselines.

We further show that semi-structured reasoning \emph{facilitates the scalable detection of reasoning flaws}.
For example, in Figure~\ref{fig:approach}, one can observe that a desired rule was skipped by comparing the number of \texttt{\small evaluate\_rule} steps with the length of the rule list returned by \texttt{\small analyze\_input}.
We refer to such checks as \emph{reasoning audits}. 
Figure~\ref{fig:approach} also provides natural-language descriptions of two audits alongside their corresponding \emph{structured reasoning audits}.
Our results indicate that these manually implemented audits can identify potential reasoning flaws and flag outcomes that are likely incorrect.

Our DSL for structured queries uses trace that has been encoded as a Pandas DataFrame, and audits also look like Python unit tests---two widely-used programming constructs. 
Because of these design choices, we show that \emph{structured reasoning audits are also easily generated automatically by modern LLMs given minimal guidance}, substantially reducing the cost of auditing reasoning in a new domain.

Beyond enabling structured reasoning queries, \emph{semi-structured reasoning facilitates additional forms of analysis}.
A recurring question in the literature~\citep{kambhampati2025reasoning} is whether reasoning LLMs generate novel ``reasoning patterns'' or simply reproduce patterns that are seen during training. 
This issue is difficult to address without a formal definition of ``reasoning patterns.''
In this work, we explore certain definitions of a ``reasoning pattern'' for semi-structured reasoners and use it to build \emph{probabilistic models of reasoning patterns for specific tasks}.  
We evaluate the hypothesis that model accuracy correlates with the probability assigned to its reasoning patterns. 
By analogy with structured audits,  we term this a \emph{typicality audit}, and show that they can also identify potential reasoning errors.

In summary, this paper makes the following contributions:
\begin{itemize}[leftmargin=*, itemsep=0pt]
\item We introduce two-stage training recipes for SSRMs that produce semi-structured reasoning traces.
\item We illustrate that both manually-generated and LLM-generated structured audits can effectively reveal potential reasoning flaws, and that failing certain audits increases the probability of error.
\item We show that typicality audits can reveal common reasoning patterns linked to better outcomes.
\item We demonstrate that auditability comes without cost in generalization performance, as SSRMs achieve results comparable to similarly trained unstructured reasoning models and other baselines.
\end{itemize}

%% file: figures/problem.tex
\begin{figure}[t]
  \centering
\includegraphics[width=\textwidth]{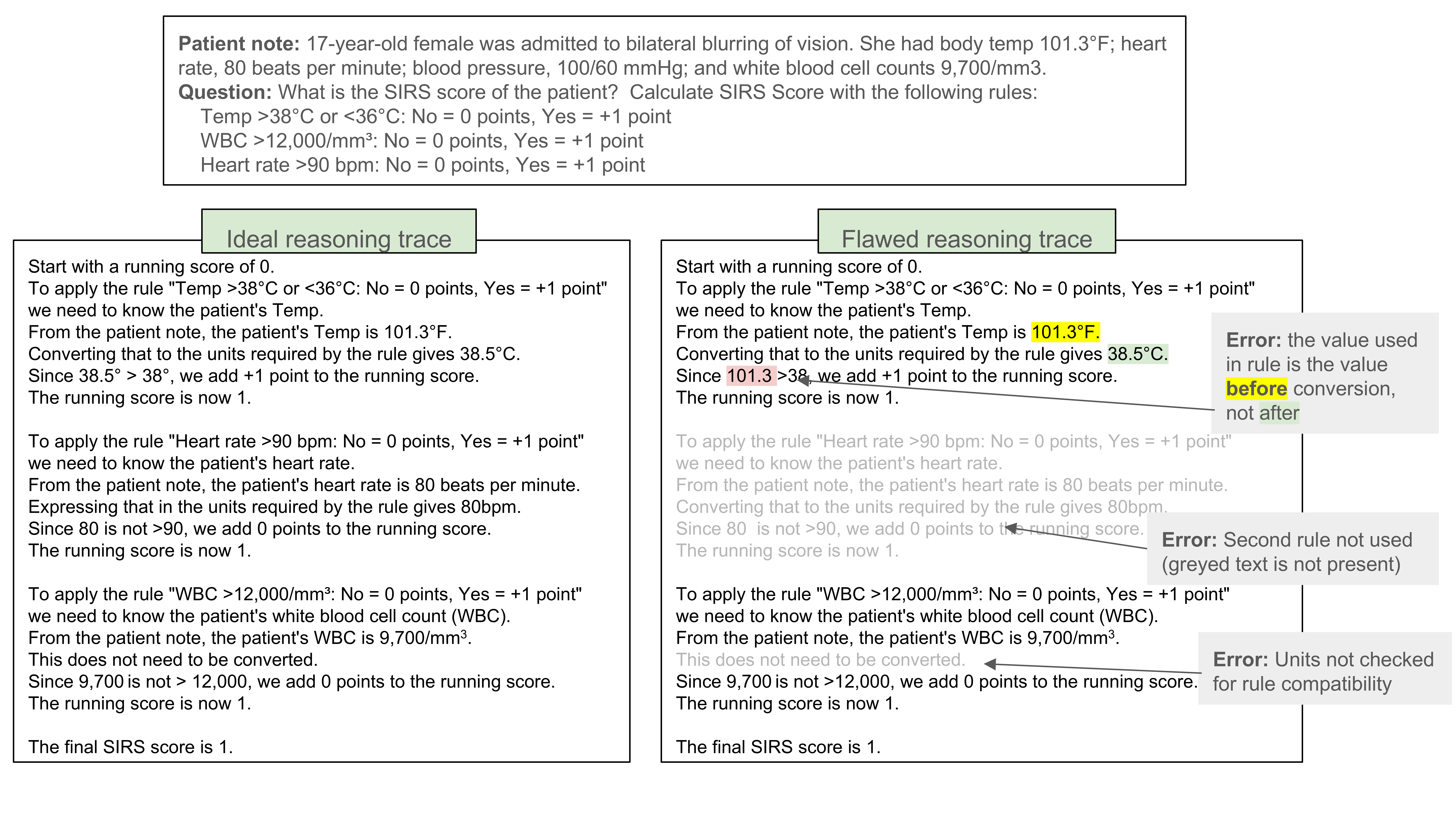}
\vspace{-1.0em}
\caption{Overview of the problem addressed. 
Top: a question that requires the LLM to extract information and apply reasoning to answer correctly.  Bottom left: a desired reasoinng trace.  Bottom right: a flawed reasoning trace.  The flawed trace differs from the desired one in three ways: (1) the incorrect patient measurement is used to determine  applicability of the first rule; (2) the second rule is skipped; (3) the units associated with a patient measurement are not explicitly checked against those required by the third rule.
In this example, none of these reasoning flaws affects the final answer, so this flawed reasoning trace will be reinforced during reinforcement learning with an outcome reward.
} 
\vspace{-1.0em}
\label{fig:problem}
\end{figure}

%% file: figures/approach.tex
\begin{figure}[t]
  \centering
\includegraphics[width=\textwidth]{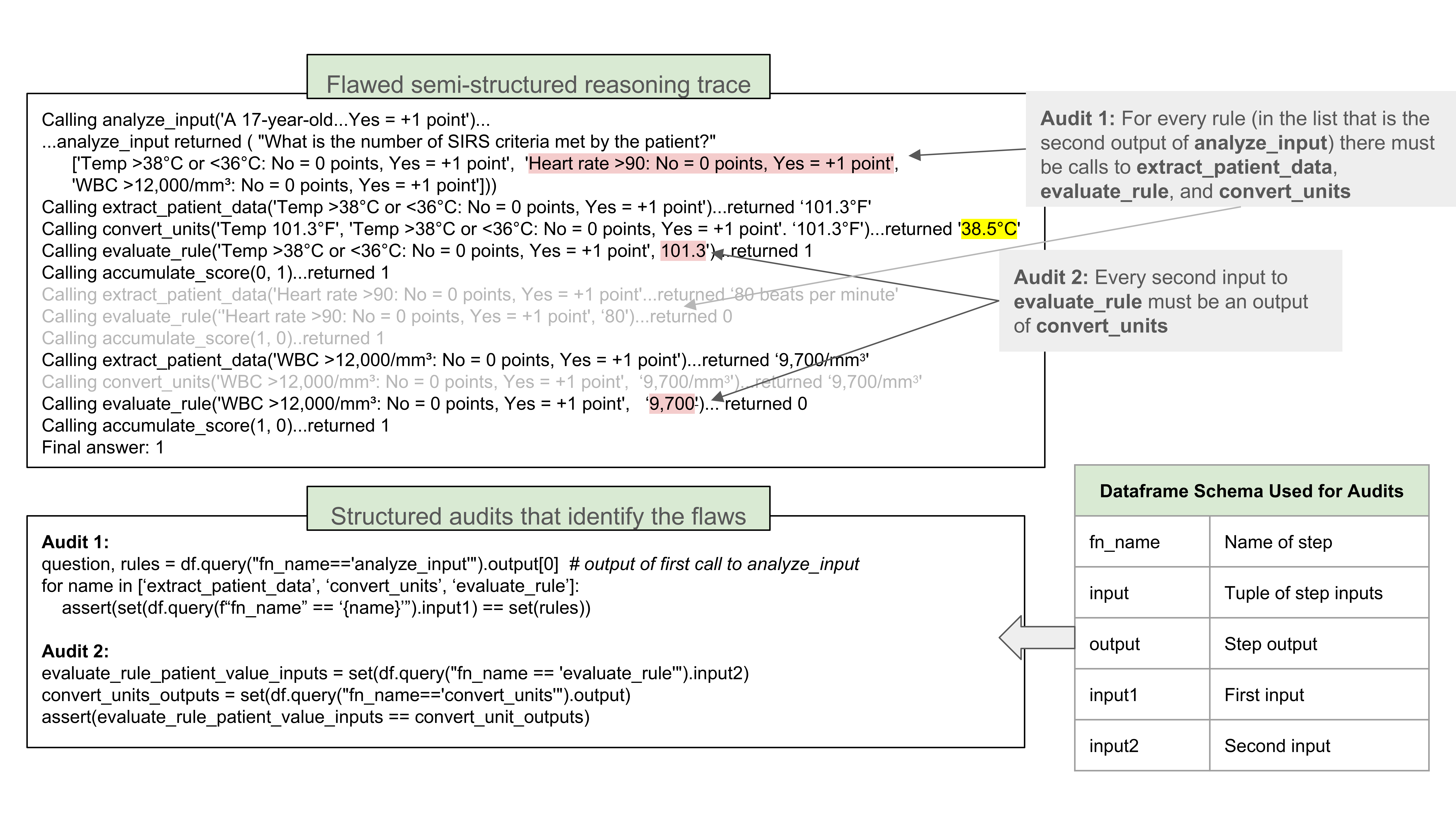}
\vspace{-1.0em}
\caption{Overview of our approach.
An LLM is trained to generate a \emph{semi-structured trace} comprising a function name, its inputs, and its outputs for each reasoning step.
Two plausible constraints on this semi-structured trace are also shown, given in natural language (gray boxes) and as executable tests (bottom left).
The executable tests are \emph{reasoning trace audits}, and in this case are hand-written.
We also explore
\emph{typicality audits}, which are learned from a corpus of reasoning traces.
} 
\vspace{-1.0em}
\label{fig:approach}
\end{figure}

%% file: sections/related_work.tex
\section{Related Work}

\textbf{Faithfulness and Process Models.}
CoT prompting has been shown to sometimes produce predictions that preserve underlying LLM biases, accompanied by explanations that obscure those biases~\citep{turpin2024language}.
This observation has motivated extensive research on explanation faithfulness in CoT prompting~\cite{jacovi-goldberg-2020-towards, turpin2024language, lanham2023measuring,bao2024llms}.
Nevertheless, defining and measuring faithfulness remains challenging, with some prior studies advocating quantitative approaches that assess mechanistic influence in neural networks through numerical metrics~\citep{parcalabescu2024measuringfaithfulnessselfconsistencynatural,bentham2024chainofthoughtunfaithfulnessdisguisedaccuracy,chen2025reasoning}. 
In this work, we propose \emph{reasoning audits} as a concrete and testable alternative to measuring faithfulness.

Other studies have proposed methods verifying reasoning chains using \emph{process reward models}~\citep{paul2024making,sun2024easy} and \emph{step reward models} \cite{viteri2024markovian,wang2023math,saparov2022language,lai2024step}.
However, these reward models are typically tailored to specific domains---such as mathematics~\citep{paul2024making,sun2024easy} or theorem-proving~\citep{saparov2022language,lai2024step}---and often rely on Monte Carlo Tree Search~\citep{kocsis2006bandit} to explore and evaluate multiple candidate reasoning chains, a computationally expensive procedure.
While our work is largely orthogonal, the symbolic and statistical audits we propose could provide complementary signals for future reward-model training. 
In particular, the statistical audits we proposed refine the notion of reasoning patterns, which have previously been identified either through task-specific analyses~\citep{zhang2025enhancing} or via LLM pipeline methods~\citep{zhou2025self}.

\textbf{Semi-structured LLM Reasoning.}
Various prompting strategies---such as CoT~\citep{wei2022chain}, Tree-of-Thought (ToT)~\citep{yao2023tree}, Chain-of-Code (CoC)~\citep{li2023chain}, and Program-of-Thought (PoT)~\citep{chen2022program}---have been widely employed to enhance the reasoning capabilities of LLMs.
More recently, research has shifted from prompting toward inference-time scaling by incorporating search algorithms, particularly tree-based search (including Monte Carlo Tree Search variants) and beam search, into the sampling process~\citep{feng2023alphazero, trinh2024solving, xin2024deepseek, kocsis2006bandit}; by ensembling multiple reasoning trajectories through self-consistency~\citep{wang2022self, huang2025efficient,aggarwal2023let}; and by applying reinforcement learning (RL) to extend the length of reasoning~\citep{o1, shao2024deepseekmath, guo2025deepseek, QwQ, kumar2024training, yang2025qwen3}.
Despite these, LLMs frequently produce reasoning traces that appear plausible yet incorrect, and such errors can be difficult to detect.

Previous studies have proposed that faithfulness can be improved by using a code-like format for LLM outputs.
Prior work assumes this format is either fully executable Python programs~\citep{chen2022program,gao2023pal,lyu2023faithfulchainofthoughtreasoning,paranjape2023art} or partially executable pseudocode~\citep{li2023chain, weir2024learning,chae2024language}. 
While enabling the use of Python as a tool often improves performance, the reasoning process used to generate the pseudocode remains obscured. 
These works have also argued (sometimes implicitly) that faithfulness is qualitatively improved with code-based outputs. 
In contrast, we pursue the more concrete goal of auditing the reasoning process.

We build most on the reasoning-chain syntax used in Program Trace Prompting (PTP)~\citep{cohen2024watch}. 
While PTP uses few-shot prompting to extrapolate ``partial programs'' and sample traces for novel inputs, SSRMs achieve strong performance without task-specific few-shot prompts.

%% file: sections/methods.tex
\section{Training Methods}
We use a two-stage training recipe for a Semi-Structured Reasoning Model (SSRM).
The first stage performs SFT to teach the model to produce the semi-structured reasoning traces, while the second stage uses reinforcement learning with verifiable rewards (RLVR) to enhance the reasoning ability.

\textbf{Supervised Fine-Tuning.}
To collect SFT data for semi-structured reasoning, we follow the PTP approach~\citep{cohen2024watch}.
We generate semi-structured reasoning traces with PTP using both Claude Sonnet 3.5 and 3.7~\citep{anthropic_claude_2024, anthropic_claude_2025} on all BBH tasks~\citep{suzgun2022challenging} as well as subsets of the training data from GSM8K~\citep{cobbe2021gsm8k}, MATH500~\citep{lightman2023let}, and MedCalcBenchV2 (please see Section~\ref{datasets}). Only traces that yield correct final answer are retained.

\textbf{Chain-of-Thought Baseline.}
To establish a fair baseline for comparison, we construct a standard CoT dataset.
We generate CoT traces on BBH using the original few-shot prompts applied to Claude Sonnet 3.5, and augment the training data with ground-truth CoT solutions from GSM8K, MATH500, and MedCalcBenchV2, for the same problem instances used in the semi-structured SFT training data.

\textbf{Training Template.}
We structure each example using a consistent markup format.
In the semi-structured setting, partial programs are wrapped in \texttt{<partial\_program>} tags, reasoning traces in \texttt{<program\_trace>} tags, both enclosed within a \texttt{<think>} tag. The final answer is placed inside the \texttt{<answer>} tag for easy parsing. For the CoT baseline, only \texttt{<think>} and \texttt{<answer>} tags are used.

\textbf{RLVR Dataset.}
In the second stage, we enhance the SFT model with RLVR.
We construct the RLVR dataset by sampling eight responses per problem from the English subset of DAPO-Math-17K~\citep{yu2025dapo}, using the SFT checkpoint.
We randomly discard half of the samples with pass rates of either zero or one.
We further include a held-out subset of MedCalcBenchV2 excluded from SFT.

\textbf{Reward Design.}
We adopt a rule-based reward combining outcome accuracy and structural validity.
Outcome accuracy measures the correctness of the final answer, while format rewards are assigned if the reasoning trace conforms to the semi-structured or CoT format, evaluated via regular expressions.

\textbf{RL Algorithm.}
We optimize with the Group Relative Policy Optimization (GRPO)~\citep{shao2024deepseekmath}, which estimates token-level advantages without requiring a critic. 
For a specific question-answer pair ($q, a$), the policy model first samples a group of $G$ individual responses $\{\mathbf{o}_i\}_{i=1}^G$.
Subsequently, the advantage of the $i$-th response is calculated as
${A}_{i, t}=\tfrac{r_i-\operatorname{mean}\left(\left\{R_i\right\}_{i=1}^G\right)}{\operatorname{std}\left(\left\{R_i\right\}_{i=1}^G\right)}$.
And the training objective is
\begin{equation}
\begin{aligned}
    \mathcal{J}_{\mathrm{GRPO}}(\theta) &= \mathbb{E}_{(q, a) \sim \mathcal{D},\left\{\mathbf{o}_i\right\}_{i=1}^G \sim \pi_{\theta_{\mathrm{old}}}(\cdot \mid q)} \\
    &\quad \left[\frac{1}{G} \sum_{i=1}^G \frac{1}{\left|\mathbf{o}_i\right|} \sum_{t=1}^{\left|\mathbf{o}_i\right|}\left(\min \left(r_{i, t}(\theta) {A}_{i, t}, \mathrm{clip}\left(r_{i, t}(\theta), 1-\varepsilon, 1+\varepsilon\right) {A}_{i, t}\right)-\beta D_{\mathrm{KL}}\left(\pi_\theta \| \pi_{\mathrm{ref}}\right)\right)\right], \\
    \text{where}\quad r_{i, t}(\theta) &= \frac{\pi_\theta\left(\mathbf{o}_{i, t} \mid q, \mathbf{o}_{i,<t}\right)}{\pi_{\theta_{\mathrm{old}}}\left(\mathbf{o}_{i, t} \mid q, \mathbf{o}_{i,<t}\right)}
\end{aligned}
\label{eq:grpo_loss}
\end{equation}
Differ from standard GRPO, we adopt fully on-policy training and token-level loss~\citep{yu2025dapo}.

%% file: sections/experiment.tex
\section{Experiments}\label{main:experiment}
In this section, we present a series of experiments conducted across diverse benchmarks---including mathematics, medical, and health domains---covering both in-domain datasets and those outside the training mixture.
We also compare SSRMs to strong prompted baselines.
Our goal is to address three key questions:
(1) Can the reasoning traces of SSRMs be audited, either through structured queries or statistical methods?
(2) Can prompted models be audited in a similar manner?
(3) Is semi-structured reasoning more difficult to learn?
Detailed setups and dataset descriptions are listed in 
Appendix~\ref{appendix_evaluation_datasets}.

\textbf{Experimental Setup.}
We use Qwen2.5-7B~\citep{yang2024qwen2} as the base model for SSRM and conduct auditability analysis on its generated semi-structured reasoning traces. 
To further validate the performance, we also train an SSRM based on Llama3.1-8B~\citep{grattafiori2024llama}.
All models are trained using \texttt{verl}~\citep{sheng2024hybridflow} on 8 H100 GPUs, with
evaluations conducted on 1 H100.

In addition to similarly trained unstructured baselines, we compare SSRMs against baselines of comparable size.
Non-reasoning baselines include Llama3.1-8B-Instruct~\citep{grattafiori2024llama}, Medical Llama~\citep{ContactDoctor} (fine-tuned for biomedical knowledge), and the Qwen series~\citep{yang2024qwen2}. 
Reasoning baselines include the DeepSeek-Distilled series~\citep{guo2025deepseek}. 
For prompted baselines, we evaluate Claude Sonnet 3.5~\citep{anthropic_claude_2024} and Qwen2.5-7B-Instruct.

We use greedy decoding and report accuracy for all tasks, except for AIME24, where we sample 32 responses and report Pass@1 with a temperature of 0.7. 
The maximum generation length is set to 32,768 tokens.
All tasks are evaluated in a zero-shot setting, except for prompted baselines, which use two-shot prompts.
Reasoning baselines follow the recommended setting (temperature 0.6, top-$p$ 0.95)~\citep{guo2025deepseek}. 
For Qwen2.5-7B, we omit the chat template following~\citet{liu2025understanding}.

\textbf{Primary Evaluation: MedCalcBenchV2.}\label{datasets}
Our primary evaluation benchmark is MedCalcBenchV2, a cleaned version of MedCalcBench~\citep{khandekar2024medcalc} (See Appendix~\ref{appendix_medcalcv2}).
MedCalcBenchV2 measures an LLM's ability to extract information from clinical text (\emph{patient note}) and perform calculations using either explicit rules or formulas provided in the prompt.
We observe that rule-based tasks are substantially more challenging than formula-based tasks.
Errors in formula-based problems primarily arise from computation or extraction mistakes, whereas errors in rule-based problems more often involve failures to follow explicit instructions, consistent with prior findings on rule-following tasks~\citep{sun2024beyond}.
To account for this discrepancy, we treat the two categories as two distinct tasks: MedCalcV2 Rules and MedCalcV2 Formulas.
Evaluation follows the original MedCalcBench criteria, which allow small numeric deviations and employ rule-based checks for date-based answers.

\textbf{Domain-Specific Language for structured audits.}
Our DSL for structured audits looks like Python unit tests: they are class methods, can be called without arguments, and contain assertion statements invoked by the class method \texttt{self.assertTrue}. 
An audit fails if it raises an exception or if an \texttt{assertTrue} call does not hold. 
The method can access a Pandas DataFrame \texttt{self.df} that represents the semi-structured trace, and assertions usually operate on this data structure using Pandas operations.

\textbf{Additional Evaluation.}
To evaluate the generalizability of the SSRMs beyond in-domain data, we conduct additional experiments on a range of benchmarks: general reasoning (GPQA-Diamond~\citep{rein2024gpqa}), mathematical reasoning (AIME24), commonsense reasoning (CommonsenseQA~\citep{talmor2018commonsenseqa}), truthfulness (TruthfulQA~\citep{lin2021truthfulqa}), as well as several medical and health-related tasks, namely MedQA~\citep{jin2020disease}, the biology and health subsets of MMLU-Pro~\citep{wang2024mmlu}, and PubMedQA~\citep{jin2019pubmedqa}, which we convert to multiple-choice.

\input{tables/struct_audits_medcalc}

\input{tables/gen_struct_audits_medcalc}

\subsection{Experimental Results}
\textbf{Both hand-crafted and LLM-generated structured audits are effective for auditing semi-structured reasoning traces generated by SSRMs.}
To validate that semi-structured reasoning can be systematically audited, we first apply hand-crafted audits for the two MedCalcV2 tasks based on the analysis of the training examples.
Table~\ref{tab:structaudit} reports results for all individual audits that are applied with sufficient frequency\footnote{Audits may not be applied to all traces---for example, one cannot confirm that number of rules evaluated is the same as the number of rules extracted if rule extraction fails to produce a legal output.} and are sufficiently discriminative---specifically, audits that succeed at least 5\% of the time and fail at least 5\% of the time.
The second audit for MedCalcV2 Formulas (e.g., ``math is correct'') uses Python's \texttt{\small eval} function; whereas all other audits inspect only trace structure.


As suggested in Figure~\ref{fig:approach}, reasoning flaws do not always yield incorrect outcomes.
In Table~\ref{tab:structaudit}, for each audit $a$, we present test accuracy when $a$ fails (``Failing'' column), when $a$ passes (``Passing'' column), the accuracy difference $\Delta$, and the statistical significance of the difference being non-zero.

The results suggest that reasoning errors are more frequent in MedCalcV2 Rules than in Formulas.
While math errors in Formulas occur frequently, they do not correlate with outcome errors.\footnote{MedCalcV2 numerical answers are soft-matched to the target, whereas the implemented audits check exact equivalence before and after simplification.} 
In contrast, reasoning errors in Rules are common and associated with substantial accuracy losses.
The most common failure is mis-summing rule contributions, followed by skipping a rule.
Other failing audits indicate mismatches between the counts of patient data extraction and rule application steps.

Because manually generating audits is expensive, we also explore automatic generation of structured audits using LLMs. 
We manually write audits for three additional tasks from BBH, and use those as few-shot examples to prompt Claude Sonnet 4.0 to output structured audits given a set of three correct sample traces.
The results, shown in Table~\ref{tab:gen-struct-audit}, show that LLM-generated structured audits are comparably useful to hand-crafted ones.  
(For more results and details, please check Appendix~\ref{appendix_additional_results})
 
\textbf{Typicality audits are also applicable for auditing semi-structured reasoning traces generated by SSRMs.}
Typicality audits provide a complementary use of the semi-structured format by analyzing \emph{abstract versions of  reasoning processes}, aka ``reasoning patterns''~\citep{zhang2025enhancing}.
Prior work has conjectured that LLMs predominantly reproduce ``reasoning patterns'' observed in the training data and struggle to generate novel sequences---i.e., LLM reasoning often relies on retrieving previously seen reasoning examples~\citep{kambhampati2025reasoning}.
If this holds, reasoning within a given task should exhibit regularity, thereby enabling statistical analyses to flag outlier traces as potential errors.

In past work, ``reasoning patterns'' are typically identified heuristically or by LLMs~\citep{zhang2025enhancing,zhou2025self}.
Here we define ``reasoning patterns'' as the sequence of step names.
For example, in Figure~\ref{fig:approach}, the pattern is ``\texttt{\small analyze\_input extract\_patient\_data convert\_units evaluate\_rule accumulate\_score extract\_patient\_data  evaluate\_rule accumulate\_score}''.
We then construct a probabilistic model $M$ over these sequences, treating them as language tokens. 
This formulation yields a precise version of the conjecture above: \emph{LLM correctness is positively correlated with the probability of the required reasoning pattern under $M$}.
To test this, we compute the correlation between outcome correctness and the probability of the reasoning pattern generated by the SSRMs.

We consider the following types of \emph{reason pattern typicality} models $M$: a \emph{unigram} language model smoothed with a Dirichlet prior (referred to as \emph{multinomial} in the tables below); \emph{bigram} and \emph{trigram} models, implemented simply by extending the base vocabulary to consider all $n$-grams of step names for $n=2,3$; an HMM with three hidden states over trigrams, denoted \emph{HMM(3,3)} in the table; and a final model, \emph{HMM$^*$}, in which we perform a grid search over different $n$-gram sizes and numbers of hidden states, selecting the configuration that optimizes the BIC criterion 
(Please see Appendix~\ref{app:audits}).


\input{tables/typicality_medcalc}
Table~\ref{tab:typicality_medcalc} summarizes the results obtained by fitting these models to the test data.\footnote{Note that the correctness label is not used in this step. Moreover, the distribution of reasoning patterns in the training data may differ for tasks with ground-truth reasoning chains.}
We use Kendall's $\tau$ to measure correlation because it makes no parametric assumptions and observe only moderate correlations ranging from 0.08 to 0.26.
As another way of testing if highly typical reasoning patterns correspond to higher accuracies than atypical ones, we sort all test predictions by pattern probability and compare the accuracy of the least probable third with that of the most probable third, excluding the middle third.
Using this method, we observe significant differences in many models.
For example, under the \emph{HMM(3,3)} model, the accuracy difference ($\Delta$) between the least and most probable tertiles is approximately 25\% for both MedCalcV2 tasks.
These results suggest that \emph{a typicality model} $M$ can approximate the behavior of a structured audit by: (a) sorting predictions by typicality, (b) splitting predictions into quantiles, and (c) interpreting the top quantile as audit-passed cases, the bottom quantile as audit-failed cases, and any other intermediate quantiles as unevaluable.
\begin{wrapfigure}{R}{0.5\textwidth}
    \vspace{-1.0em}
    \centering
    \includegraphics[width=1.0\linewidth]{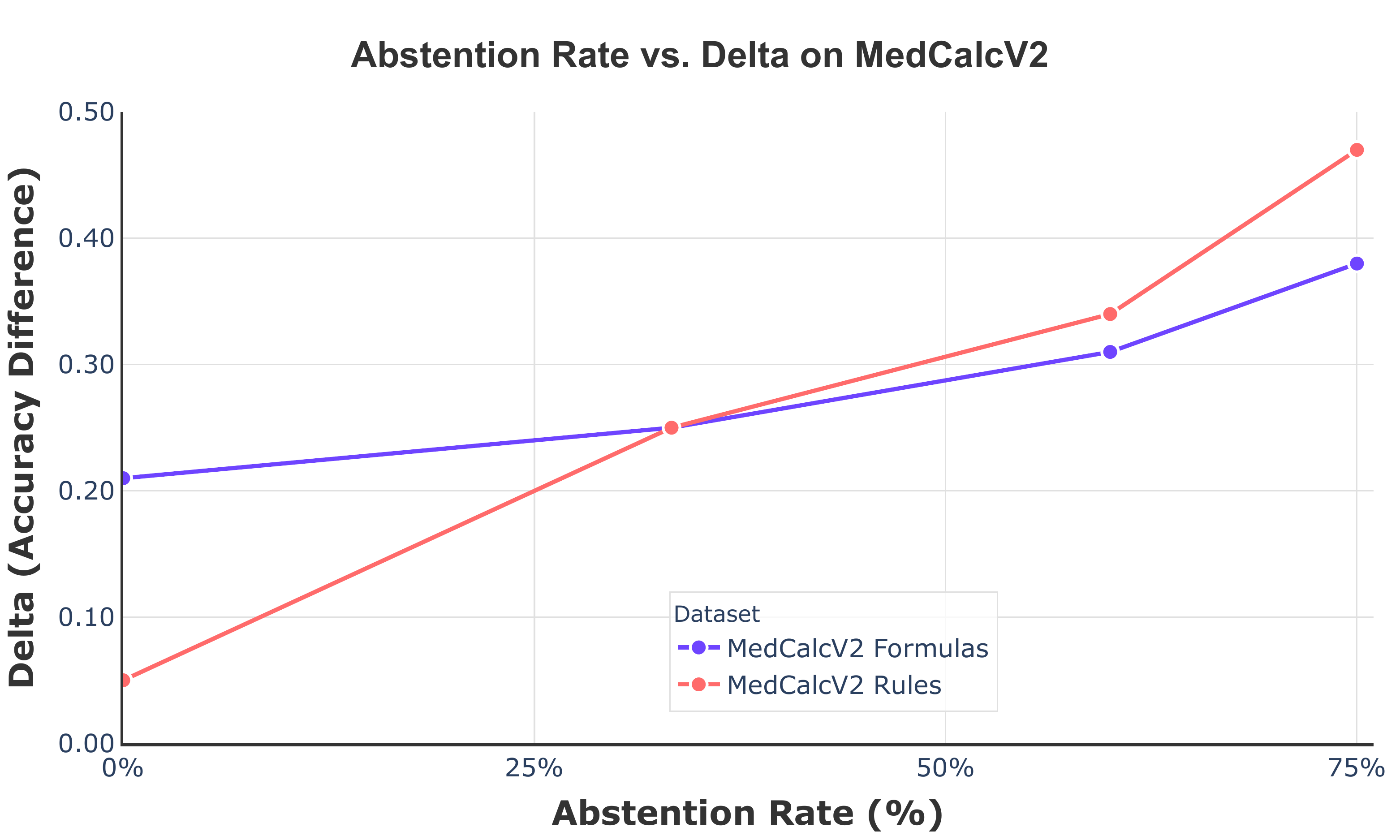}
    \vspace{-1.5em}
    \caption{Abstention Rate vs. $\Delta$ (accuracy difference between the highest- and lowest- probability quantiles) on MedCalcV2 under a typicality audit.}
    \vspace{-1.0em}
    \label{fig:abstentions}
\end{wrapfigure}
Following this approach, the typicality audits function as an \emph{abstaining classifier}~\cite{pietraszek2005optimizing} for evaluating outcome correctness, abstaining specifically on intermediate scores.
The abstention rate can be adjusted by splitting the predictions into different numbers of quantiles: for instance, dividing predictions into two groups yields no abstentions, whereas dividing them into eight groups results in abstention for the middle six octiles---i.e.,\protect{~$\frac{6}{8}=\frac{3}{4}$} of the time.
As shown in Figure~\ref{fig:abstentions}, higher abstention rates are associated with larger 
accuracy difference ($\Delta$) across both MedCalcV2 tasks.

Given the effectiveness of both structured and typicality audits in identifying potential reasoning errors, a straightforward extension is to apply apply at inference time.
For example, when combining typicality audits with self-consistency, reasoning traces in the lowest tertile can be resampled more extensively, whereas those in the highest tertile---more likely to be correct---might require no additional sampling.
This strategy can help concentrate the sampling budget on the most error-prone cases.
We evaluate this approach on the MedCalcV2 Rules tasks and report the results 
in Appendix~\ref{appendix-test-time-scaling}.
Our findings show that audit-guided self-consistency reduces computational cost while maintaining comparable or slightly improved performance relative to vanilla self-consistency with a fixed sampling budget.
However, we do not observe significant improvements over greedy decoding with a single sample.
We hypothesize that traces flagged as incorrect by audits may correspond to problems that the model struggles to solve even with additional sampling budget.
We leave this to future work.


\input{tables/struct_audit_medcalc_prompted}

\input{tables/typicality_medcalc_prompted}

\textbf{Both structured and typicality audits can be applied to semi-structured traces from few-shot prompted models.}
Table~\ref{tab:structaudit_prompt} presents the results of structured audits applied to few-shot prompted Claude Sonnet 3.5 on both MedCalcV2 tasks, while Table~\ref{tab:typicality_medcalc_prompted} shows typicality audit results for the same model (limited to three representative typicality models for brevity).
Overall, Claude Sonnet 3.5 behaves similarly to SSRMs under these audits, except for the typicality audit for Rules, where more typical reasoning traces exhibit \emph{higher} error rates on average (although not significantly so).
This may be attributed to the high error rate of the prompted model itself: even typical reasoning processes often lead to errors.
In Appendix~\ref{appendix_additional_results},
we present results for Qwen2.5-7B-Instruct---a weaker prompted model and the instruction-tuned version of the Qwen2.5-7B base used as the backbone for SSRMs---as well as LLM-generated structured audit results on prompted Claude.





\begin{wrapfigure}{R}{0.5\linewidth}
    \vspace{-1.4em}
    \centering
    \includegraphics[width=1.0\linewidth]{figures/model_comparison_prompted.png}
    \vspace{-1.8em}
    \caption{Comparison of SSRM against similarly trained and prompted baselines on MedCalcV2.}
    \label{fig:prompted}
    \vspace{-1.5em}
\end{wrapfigure}
\textbf{Semi-structured reasoning is learnable and achieves performance and generalization on par with unstructured reasoning.}
We consider two different model families for training SSRMs: a stronger one based on Qwen2.5-7B and a weaker one based on Llama3.1-8B.
Our trained Qwen SSRM achieves strong results, as shown in Table~\ref{tab:bigtable}.
On average, \emph{it outperforms the unstructured reasoning baseline trained with the same procedure}.
On the two challenging MedCalcV2 tasks, it exceeds six other strong baselines of comparable size.
Within the training mixture, it outperforms the best baseline by nearly ten points.
Moreover, it generalizes effectively to tasks outside the training mixture: while it does not match the top math-specialized reasoning models, it outperforms all non-reasoning baselines.
On a range of medical QA benchmarks, it achieves performance comparable to reasoning models, lagging only slightly behind BioMedical-Llama-3-8B, a specialized model for biomedical knowledge. 
By contrast, although the Llama SSRM is based on a weaker backbone, it delivers performance comparable to similarly trained unstructured reasoning baselines on both in-domain and out-of-domain benchmarks, further supporting that semi-structured reasoning does not compromise performance.
\input{tables/big_table}

In Figure~\ref{fig:prompted}, we show that Qwen SSRM not only outperforms the Claude Sonnet 3.5, which is used to seed the SFT training data, but also significantly outperforms the Qwen instruction-tuned variant.
For Claude Sonnet 3.5 and Qwen2.5-7B-Instruct, we employ two-shot prompting, using two fixed demonstrations across all MedCalc ``calculators''.\footnote{This also diverges from the MedCalcBench few-shot evaluation, which selects a single demonstration from the same calculator as the test instance.}
For each prompted model, we evaluate two prompt variants: one with unstructured free-form CoT prompts and one with the semi-structured format.

We also analyze the token usage of Qwen SSRM and unstructured reasoning baselines (see Appendix~\ref{appendix_token_usage} for details).
In summary, SSRM consume more tokens than the unstructured reasoning baselines on MedCalcV2 Tasks, while token usage is comparable on MATH500 and GPQA-D. 
One factor contributing to the increased usage is redundant argument and variable referencing, as shown in Figure~\ref{fig:approach}.
We leave the development of a more efficient referencing mechanism to future work.

%% file: tables/struct_audits_medcalc.tex
\begin{table}[htbp]

\caption{Hand-crafted structured audits for Qwen SSRM generated semi-structured traces on two MedCalcV2 tasks.
For each, we report the failure rate, the outcome accuracy conditioned on audit failing or passing, the accuracy difference ($\Delta$) between passing and failing cases, and the $p$-value for testing $\Delta\not=0$. 
One star (*) for statistical significance at $p<0.1$ and two stars (**) for $p<0.05$.}\label{tab:structaudit}
 \vspace{-0.5em}

\begin{center}
\resizebox{\linewidth}{!}{
\begin{tabular}{lrrrrrl}
\toprule
 & & \multicolumn{3}{c}{$-$ accuracy and difference $-$} \\
&  \%Failed & Failing & Passing & $\Delta$ & $p$-val & description of audit \\
\midrule
\uu{MedCalcV2 Formulas}
 & 22.0 & 0.77 & 0.86 & 0.09 & & \texttt{solve\_formula} output is formatted well \\
& 49.0 & 0.84 & 0.83 & -0.01 &  & \texttt{solve\_formula} math is correct$^\textrm{math}$  \\
\midrule
\uu{MedCalcV2 Rules}
& 13.2 & 0.22 & 0.46 & 0.24 & ** & one \texttt{get\_data} step per extracted rule \\
&13.4 & 0.22 & 0.47 & 0.25 & ** & \texttt{get\_data} called on all rules \\
& 14.0 & 0.21 & 0.47 & 0.26 & ** & one \texttt{eval\_rule} step per rule \\
& 20.3 & 0.26 & 0.48 & 0.22 & ** & all rule outputs summed correctly \\

\bottomrule

\end{tabular}
}

\end{center}
\vspace{-1.5em}
\end{table}

%% file: tables/gen_struct_audits_medcalc.tex
\begin{table}[htbp]

 \vspace{1em}

\caption{LLM-generated structured audits on the same set of Qwen SSRM traces for MedCalcV2.} \label{tab:gen-struct-audit}

 \vspace{-0.5em}

\begin{center}
\resizebox{\linewidth}{!}{
\begin{tabular}{lrrrrrl}
\toprule
 & & \multicolumn{3}{c}{$-$ accuracy and difference $-$} \\
&  \%Failed & Failing & Passing & $\Delta$ & $p$-val & description of audit \\
\midrule
\uu{MedCalcV2 Formulas}

& 5.7 & 0.76 & 0.84 & 0.08 &  & step 4 output feeds into step 5 input \\
& 6.4 & 0.45 & 0.86 & 0.41 & ** & step 3 output feeds into step 4 input \\
& 8.3 & 0.62 & 0.86 & 0.24 &  & sstep 2 output feeds into step 3 input \\
& 9.3 & 0.82 & 0.84 & 0.02 &  & convert\_units called once per datapoint \\
& 10.2 & 0.81 & 0.84 & 0.03 &  & convert\_units receives formula as first input \\
& 12.6 & 0.80 & 0.84 & 0.05 &  & convert\_units correct second input \\
& 14.4 & 0.37 & 0.92 & 0.55 & ** & get\_data receives formula from analyze\_input \\
\midrule
\uu{MedCalcV2 Rules}
 & 13.4 & 0.22 & 0.47 & 0.25 & ** & get\_data called for each rule \\
 & 13.9 & 0.21 & 0.47 & 0.26 & ** & consistent rules across get\_data steps \\

\bottomrule

\end{tabular}
}

\end{center}
\vspace{-1.5em}
\end{table}

%% file: tables/typicality_medcalc.tex
\begin{wraptable}{R}{0.5\linewidth}
\vspace{-1.2em}
  \centering
    \caption{The results prove that atypical reasoning pattern in the MedCalcV2 tasks are more likely to result in errors.
    We evaluate several typicality/probability models, all of which correlate with correctness, though the correlation is weaker on MedCalcV2 Rules.
    In additional to Kendall's $\tau$ for correlation, we also partition the test data into three equal groups by probability and report accuracy in the lowest and highest tertiles, the accuracy difference ($\Delta$), and the $p$-value of this difference.}  \label{tab:typicality_medcalc}
\resizebox{1.0\linewidth}{!}{
\begin{tabular}{p{2.3cm}ccccc}
\toprule
\uu{MedCalcV2 Formulas} & & \multicolumn{3}{c}{$-$ accuracy and difference $-$} & \\%
&	$\tau$ & Tertile 1 & Tertile 3 &	$\Delta$ &	$p$-val \\
multinomial &	0.25 &	0.72 &	0.95 &	0.22 &	* \\
bigram &	0.25 &	0.72 &	0.95 &	0.23 &	* \\
trigram &	0.26 &	0.72 &	0.95 &	0.23 &	* \\
HMM(3,3) &	0.26 &	0.72 &	0.97 &	\bb{0.25} &	** \\
HMM$^*$   &	0.21 &	0.74 &	0.97 &	0.07 &	   \\
\midrule
\uu{MedCalcV2 Rules} & & \multicolumn{3}{c}{$-$ accuracy and difference $-$} &\\%
&	$\tau$ & Tertile 1 & Tertile 3 &	$\Delta$ &	$p$-val \\
multinomial &	0.17 &	0.32 &	0.57 &	\bb{0.25} &	** \\
bigram &	0.17 &	0.32 &	0.57 &	\bb{0.25} &	** \\
trigram &	0.17 &	0.32 &	0.57 &	\bb{0.25} &	** \\
HMM(3,3) &	0.17 &	0.32 &	0.57 &	\bb{0.25} &	** \\
HMM$^*$    &	0.08 &	0.43 &	0.52 &	0.09 &	 \\
\bottomrule
\end{tabular}
}

\end{wraptable}

%% file: tables/struct_audit_medcalc_prompted.tex
\begin{table}[tb]

\caption{Results of applying structured audits to Claude Sonnet 3.5 with semi-structured prompting on both MedCalcV2 Formulas and Rules. 
Overall, the results resemble those of SSRM, although the prompted MedCalcV2 Formulas system does have some rarely-failing audits that impact accuracy.}\label{tab:structaudit_prompt}
\vspace{-0.8em}
\begin{center}
\resizebox{\linewidth}{!}{
\begin{tabular}{lrrrrrl}
\toprule
 & & \multicolumn{3}{c}{$-$ accuracy and difference $-$} \\
&  \%Failed & Failing & Passing & $\Delta$ & $p$-val & description of audit \\
\midrule
\uu{MedCalcV2 Formulas}
& 1.7 & 0.000 & 0.662 & 0.662 &  & one \texttt{get\_data}step \\
& 2.1 & 0.000 & 0.664 & 0.664 & * & one \texttt{insert\_variables} step \\
~~Claude Sonnet 3.5
& 3.8 & 0.091 & 0.673 & 0.582 & ** & \texttt{solve\_formula} output is a number$^\textrm{math}$ \\
~~(65.1\% acc)
 & 9.2 & 0.593 & 0.657 & 0.064 &  & \texttt{solve\_formula} output is formatted correctly \\
 & 47.3 & 0.667 & 0.636 & -0.030 &  & \texttt{solve\_formula} math is correct$^\textrm{math}$ \\
\midrule
\uu{MedCalcV2 Rules} 
& 5.8 & 0.182 & 0.399 & 0.218 &  & analyze\_input returns correct \# values \\
 & 14.7 & 0.196 & 0.420 & 0.223 & ** & one  \texttt{convert\_units} step per rule\\
~~Claude Sonnet 3.5 
 & 14.7 & 0.196 & 0.420 & 0.223 & ** & one  \texttt{get\_data} step per rule  \\
~~(38.7\% acc)
 & 15.8 & 0.183 & 0.425 & 0.242 & ** & one  \texttt{evaluate\_rule} step per rule\\
 & 17.1 & 0.169 & 0.432 & 0.263 & ** & one  \texttt{accumulate\_score} step per rule\\
\bottomrule

\end{tabular}
}

\vspace{-1.0em}
\end{center}

\end{table}

%% file: tables/typicality_medcalc_prompted.tex
\begin{wraptable}{R}{0.5\linewidth}
\vspace{-1.2em}
\centering
\caption{Results of applying typicality audits to semi-structured reasoning traces from few-shot prompted Claude Sonnet 3.5 on both MedCalcV2 Formulas and Rules.
Overall, the results indicate that the hypothesis---that atypical reasoning patterns correspond to higher error rates---holds for MedCalcV2 Formulas but not for the noisier Rules.}
\label{tab:typicality_medcalc_prompted}
\resizebox{1.0\linewidth}{!}{
\begin{tabular}{p{2.3cm}ccccc}
\toprule
 & & \multicolumn{3}{c}{$-$ accuracy and difference $-$} & \\
 & $\tau$ & Tertile 1 & Tertile 3 & $\Delta$ & $p$-val \\
\midrule
\uu{MedCalcV2 Formulas} \\
\addlinespace[2pt]
\multicolumn{6}{l}{Claude Sonnet 3.5 (65.1\% acc)} \\
\addlinespace[2pt]
trigram   & 0.13 & 0.56 & 0.67 & 0.11 &   \\
HMM(3,3)  & 0.21 & 0.56 & 0.70 & 0.14 &   \\
HMM$^*$   & 0.30 & 0.54 & 0.87 & 0.33 & **\\
\midrule
\uu{MedCalcV2 Rules} \\
\addlinespace[2pt]
\multicolumn{6}{l}{Claude Sonnet 3.5 (38.7\% acc)} \\
\addlinespace[2pt]
trigram   & -0.06 & 0.47 & 0.33 & -0.14 &   \\
HMM(3,3)  & -0.06 & 0.46 & 0.32 & -0.14 &   \\
HMM$^*$   & -0.05 & 0.43 & 0.33 &  0.00 &   \\
\bottomrule
\end{tabular}
}
\vspace{-1.0em}
\end{wraptable}

%% file: tables/big_table.tex
\begin{table}[!htb]
  \centering
  \vspace{-0.5em}
  \caption{Our models are initialized from Qwen2.5-7B and Llama3.1-8B, trained with SFT followed by RLVR on a mix of MedCalcV2 and other math tasks. 
  Underlined results indicate the best performance among our comparably trained models; starred results denote best among non-reasoning models; and bold results are best overall.
  On average, SSRM outperforms the unstructured CoT format and six strong, comparably sized baseline models.\protect\footnotemark[4] We sample 32 times and report Pass@1 for AIME24.}
  \label{tab:bigtable}
  \renewcommand{\arraystretch}{1.08}
  \setlength{\tabcolsep}{4.5pt}
  \resizebox{\linewidth}{!}{%
  \begin{tabular}{lrrrrr|rrrr|rrr|rrr}
    \toprule
    & \multicolumn{5}{c}{\makecell{SSRM from Qwen2.5-7B (ours)}} 
    & \multicolumn{4}{c}{\makecell{SSRM from Llama3.1-8B (ours)}} 
    & \multicolumn{3}{c}{\makecell{Instr/reasoning LLMs\\(Qwen2.5-7B)}} 
    & \multicolumn{3}{c}{\makecell{Instr/reasoning LLMs\\(Llama3/3.1-8B)}} \\
    \cmidrule(lr){2-6}\cmidrule(lr){7-10}\cmidrule(lr){11-13}\cmidrule(lr){14-16}
    & \makecell{Base}
    & \makecell{unstr.\\+SFT} & \makecell{unstr.\\++RL}
    & \makecell{semi-str.\\+SFT} & \makecell{semi-str.\\++RL}
    & \makecell{unstr.\\+SFT} & \makecell{unstr.\\++RL}
    & \makecell{semi-str.\\+SFT} & \makecell{semi-str.\\++RL}
    & \makecell{Instr} & \makecell{$^*$OpR1} & \makecell{$^*$DSeek}
    & \makecell{BioL} & \makecell{Instr} & \makecell{$^*$DSeek} \\
    \midrule
MedCalcV2 Formulas      & 3.0 & 52.4 & 75.9 & 63.3 & $^*$\uu{\bb{83.6}} & 48.7 & 58.9 & 56.9 & 75.0 & 56.4 & 44.8 & 36.9 & 12.6 & 10.0 & 29.7 \\
MedCalcV2 Rules         & 0.0 & 27.4 & 38.4 & 38.9 & $^*$\uu{\bb{43.2}} & 27.9 & 28.4 & 20.8 & 36.3 & 32.1 & 22.6 & 14.2 & 16.6 & 9.5  & 9.2  \\
GSM8k                   & 85.4& 74.4 & 90.5 & 76.6 & $^*$\uu{90.9}     & 42.3 & 43.8 & 57.4 & 36.2 & $^*$90.9 & \bb{94.8} & 89.2 & 51.9 & 81.1 & 75.7 \\
MATH500                 & 69.2& 44.6 & \uu{77.0} & 45.4 & 75.2          & 15.6 & 15 & 22.6 & 12.4 & $^*$\bb{78.8} & 91.0 & 94.0 & 17.4 & 46.2 & 87.2 \\
\textit{train mix avg}  & \ii{39.4} & \ii{49.7} & {\ii{70.5}} & \ii{56.1} & $^*$\uu{\bb{\ii{73.2}}} & 33.6 & 36.5 & 39.4 & 40 & \ii{64.6} & \ii{63.3} & \ii{58.6} & \ii{24.6} & \ii{36.7} & \ii{50.5} \\
\midrule
AIME24                  & 9.1  & 1.1  & 12.1 & 3.7  & $^*$\uu{12.4}     & 0.3 & 0.5 & 0.2 & 0.9 & 11.8 & 45.3 & \bb{53.4} & 0.1  & 2.1  & 45.2 \\
GPQA-D                  & 31.8 & 31.8 & $^*$\uu{38.4} & 30.3 & 34.3      & 25.8 & 33.8 & 22.2 & 26.3 & 32.8 & 41.4 & \bb{50.5} & 26.8 & 31.8 & 43.9 \\
TruthfulQA              & 49.7  & \uu{57.3} & 56.3 & 41.1 & 54.3              & 41.6 & 39.8 & 31.3 & 32.9 &  $^*$\bf{{55.6}}  &  42.6  &  47.5  &  53.0  &  54.3  &  52.6  \\
CommonsenseQA           & 70.5  & 70.1 & 72.8 & 70.8 & $^*$\uu{\bf{75.7}}              & 52.7 & 52.4 & 58.6 & 60.4 &  66.8  &  54.0  &  52.3  &  39.3  &  50.4  &  63.1  \\
\midrule
MedQA                   & 57.4& \uu{62.4} & 62.0 & 55.9 & 61.4          & 55.4 & 57 & 50.4 & 53 & $^*$62.8 & 31.1 & 36.4 & \bb{76.9} & 68.9 & 58.1 \\
MMLU Pro Bio            & 64.6& 68.6 & 71.8 & 59.3 & \uu{69.9}         & 58 & 58.4 & 55.4 & 59.3 & $^*$\bb{73.5} & 50.9 & 66.7 & 64.6 & 67.8 & 73.1 \\
MMLU Pro Health         & 42.1& \uu{53.2} & 53.1 & 40.5 & 51.7          & 40.1 & 41.6 & 40.1 & 39.7 & $^*$54.8 & 22.0 & 33.4 & 53.1 & \bb{58.3} & 46.5 \\
PubmedQA                & 66.3& 73.4 & 71.4 & 70.2 & $^*$\uu{76.2}     & 73.9 & 75.5 & 68.2 & 73.4 & 73.5 & 73.3 & 72.7 & \bb{77.1} & 75.6 & 73.8 \\
\textit{med/health avg} & \ii{57.6} & \ii{64.4} & \ii{64.6} & \ii{56.5} & \uu{\ii{64.8}} & 56.9 & 58.1 & 53.5 & 56.4 & \ii{66.2} & \ii{44.3} & \ii{52.3} & $^*$\bf{\ii{67.9}} & \ii{67.7} & \ii{62.9} \\
\midrule
\textit{overall avg}    & \ii{45.3} & \ii{51.3} & \ii{60.8} & \ii{50.2} & $^*$\uu{\bb{\ii{61.7}}} & \ii{39.7} & \ii{41.7} & \ii{40.3} & \ii{42} & \ii{58.0} & \ii{52.1} & \ii{54.3} & \ii{39.5} & \ii{45.6} & \ii{54.6} \\
\bottomrule
  \end{tabular}}
  \footnote\ {\small All accuracies are percentages. ``Instr" models are instruction-trained, ``DSeek" are distilled from DeepSeek-R1, and OpR1 is OpenR1-Qwen-7B. BioL is Bio-Medical-Llama-3-8B.}
\end{table}

%% file: sections/conclusion.tex
\section{Conclusion}

We have presented methods for scalably testing whether an LLM adheres to a prescribed reasoning strategy on specific critical tasks.
Our methods combine a Semi-Structured Reasoning Model (SSRM), which outputs reasoning steps in a semi-structured format, with methods for \emph{auditing} these reasoning traces.
We consider two challenging tasks: (a) extracting information from clinical text and (b) performing a series of calculations using the extracted values, based on either predefined rules or given formulas.
These tasks are adapted from MedCalcBench, which has been cleaned, deduplicated, and restructured to separate the simpler formula-based tasks from the more complex rule-based ones.

We show that \emph{structured reasoning audits} can reveal meaningful classes of likely reasoning errors for these tasks and qualitatively distinguish between the types of errors made across tasks and models.
We further introduce \emph{typicality audits}, which are probabilistic models trained on a corpus of semi-structured reasoning traces.
Typicality audits approximate structured audits by (a) sorting predictions by typicality, (b) splitting predictions into quantiles, and (c) interpreting the top quantile as a pass and the bottom quantile as a fail. 
Both types of audits can be applied to few-shot prompted models.

Importantly, auditability appears to come without a cost in accuracy: overall, our Qwen SSRM model outperforms plausible baselines, including strong closed-source prompted models, an identically-trained unstructured baseline, and many other strong comparably-sized models. Likewise, the Llama SSRM demonstrates comparable performance relative to its identically-trained unstructured baseline.

%% file: sections/appendix.tex

\section{Limitations}
We demonstrate SSRMs' effectiveness on Qwen2.5-7B and Llama3.1-8B. 
Experiments with different architectures and extend to larger scales could help clarify the generalizability of the technique.

While symbolic audits provide a novel mechanism for monitoring behavior of LLMs, they can only capture some aspects of intended behavior. If audit coverage is incomplete, a model might pass all audits while following a logically incorrect reasoning process.  (This limitation is analogous to the use of unit tests in software development, where test coverage is often incomplete).
Additionally, models can execute individual steps incorrectly---a failure mode that reasoning audits typically fail to detect.

Typicality audits identify reasoning traces that are unusual, which need not be correlated with traces that are incorrect (e.g., if a model has a high error rate, highly typical traces might still be incorrect.)


In this study, we conducted only preliminary experiments integrating test-time-scaling with audits.
Further investigations into effectively combining audits with test-time-scaling methods---such as audit-based self-consistency---to show their utility during inference time are left for future work.

\section{Broader Impacts}

This paper introduces Semi-Structured Reasoning Models (SSRMs) and presents two types of audits to identify probable reasoning errors in the semi-structured reasoning traces: (1) hand-crafted or LLM-generated structured audits and (2) probabilistic model-based typicality audits. 
Our goal is to detect undesirable reasoning shortcuts for LLMs while maintaining good downstream performance.

\section{Training Details}\label{appendix_training_details}
Detailed hyperparameters configurations for both Stage 1 (SFT) and Stage 2 (RLVR) are provided in Table~\ref{tab:appendix_hyperparameters}.
We provide the detailed settings in subsequent subsections to support reproducibility.

\input{tables/hyperparameters}
\subsection{Supervised Fine-Tuning (SFT) Configurations}
Table~\ref{appendix_sft_system_prompt} presents the system prompt template we used for SSRMs.
The same system prompt is used for both Stage 1 and 2.
Table~\ref{appendix_training_sample_demo} shows a semi-structured reasoning trace from GSM8K used for SFT.

\begin{figure*}[htbp]
\scriptsize
\centering
\begin{tcolorbox}[colback=white, colframe=customBlue, width=1.0\textwidth, arc=3mm, boxrule=0.5mm, title=System Prompt Template for Semi-Structured Reasoning Models (SSRMs)]
\begin{Verbatim}[breaklines=true, breakanywhere=true, formatcom=\bfseries]
A conversation between User and Assistant. The User asks a question, and the Assistant solves it. The assistant first reasons through the problem by generating high-level partial programs with key parts hidden using "..." markers. It then simulates programs trace based on the incomplete partial programs. The partial program must be general enough to solve all instances of the problem type, not just specific examples. The partial programs and traces are enclosed within <partial_program> </partial_program> and <program_trace> </program_trace> tags, while the overall reasoning process and final answer are enclosed within <think> </think> and <answer> </answer> tags, respectively. You should also wrap your final answer in $\\boxed{{ANSWER}}$ if it is a mathematical expression.

Format:
<think>
<partial_program>
[Partial Program here]
</partial_program>
<program_trace>
[Program Trace here]
</program_trace>
</think>
<answer>
[Final Answer here]
</answer>
\end{Verbatim}

\end{tcolorbox}
\caption{System Prompt Template for Semi-Structured Reasoning Models (SSRMs).}
\label{appendix_sft_system_prompt}
\end{figure*}

\begin{figure*}[htbp]
\scriptsize
\centering
\begin{tcolorbox}[colback=white, colframe=customBlue, width=1.0\textwidth, arc=3mm, boxrule=0.5mm, title=Semi-Structured Reasoning Trace for GSM8K]
\begin{Verbatim}[breaklines=true, breakanywhere=true, formatcom=\bfseries]
<think>
<partial_program>
@traced
def analyze_input(input_str: str) -> tuple[str, ...]:
 """From an input, extract a tuple of individual questions and given values, as well as the final question to be answered.
 """
 ...

@traced
def convert_to_equations(input_questions: tuple[str, ...]) -> tuple[str, ...]:
 """Takes a list of questions and converts them to a list of mathematical equations.
 """
 ...

@traced
def simplify_equation(input_equation: str, given_values: tuple[str, ...]) -> str:
 """Takes an input equation and simplifies it by replacing a variable with the value of the variable as found in a list of given values if possible or performing algebraic operations if not, returning the simplified equation.
 """
 ...

</partial_program>

<program_trace>
Calling analyze_input("Two-fifths of the seats in an auditorium that holds 500 people are currently taken. It was found that 1/10 of the seats are broken. How many seats are still available?")...
...analyze_input returned ('An auditorium holds 500 people.', 'Two-fifths of the seats are currently taken.', '1/10 of the seats are broken.', 'How many seats are still available?')
Calling convert_to_equations(('An auditorium holds 500 people.', 'Two-fifths of the seats are currently taken.', '1/10 of the seats are broken.', 'How many seats are still available?'))...
...convert_to_equations returned ('total_seats = 500', 'occupied_seats = total_seats * 2/5', 'broken_seats = total_seats * 1/10', 'available_seats = total_seats - occupied_seats - broken_seats')
Calling simplify_equation('available_seats = total_seats - occupied_seats - broken_seats', ('total_seats = 500', 'occupied_seats = total_seats * 2/5', 'broken_seats = total_seats * 1/10'))...
...simplify_equation returned 'available_seats = 500 - occupied_seats - broken_seats'
Calling simplify_equation('available_seats = 500 - occupied_seats - broken_seats', ('total_seats = 500', 'occupied_seats = total_seats * 2/5', 'broken_seats = total_seats * 1/10'))...
...simplify_equation returned 'available_seats = 500 - (500 * 2/5) - broken_seats'
Calling simplify_equation('available_seats = 500 - (500 * 2/5) - broken_seats', ('total_seats = 500', 'occupied_seats = total_seats * 2/5', 'broken_seats = total_seats * 1/10'))...
...simplify_equation returned 'available_seats = 500 - 200.0 - broken_seats'
Calling simplify_equation('available_seats = 500 - 200.0 - broken_seats', ('total_seats = 500', 'occupied_seats = total_seats * 2/5', 'broken_seats = total_seats * 1/10'))...
...simplify_equation returned 'available_seats = 500 - 200.0 - (500 * 1/10)'
Calling simplify_equation('available_seats = 500 - 200.0 - (500 * 1/10)', ('total_seats = 500', 'occupied_seats = total_seats * 2/5', 'broken_seats = total_seats * 1/10'))...
...simplify_equation returned 'available_seats = 500 - 200.0 - 50.0'
Calling simplify_equation('available_seats = 500 - 200.0 - 50.0', ('total_seats = 500', 'occupied_seats = total_seats * 2/5', 'broken_seats = total_seats * 1/10'))...
...simplify_equation returned 'available_seats = 250.0'
</program_trace>
</think>
<answer>
250
</answer>
\end{Verbatim}

\end{tcolorbox}
\caption{Semi-Structured Reasoning Trace for GSM8K.}
\label{appendix_training_sample_demo}
\end{figure*}

\subsection{Reinforcement Learning with Verifiable Rewards (RLVR) Configurations}
\textbf{Reward Design.}
We employ two types of rule-based rewards functions for reinforcement learning:
\begin{itemize}[leftmargin=*, itemsep=0pt, topsep=0pt]
\item \textbf{Outcome Rewards:} The generated response will be assigned a reward of 1 for correct answers and 0 for incorrect answers. 
For the MedCalcV2 data, we follow the original proposed protocol and allow for a range-based evaluation; for the DAPO math data, we use an exact-match criterion.

\item \textbf{Format Rewards:} We require all models to format its responses using tags such as \texttt{<think>} and \texttt{<answer>}.  For SSRMs specifically, additionally require the tags \texttt{<partial\_program>} and \texttt{<program\_trace>}, define at least three functions within the \texttt{<partial\_program>} block, and exclusively invoke these functions within the \texttt{<program\_trace>} block. 
Given the regular syntax of semi-structured reasoning traces, these constraints can be enforced via regular expressions.
\end{itemize}

\section{Experimental Details}\label{appendix_evaluation_datasets}
\subsection{MedCalcBench V2}\label{appendix_medcalcv2}
The original MedCalcBench~\citep{khandekar2024medcalc} contains examples from 55 distinct \emph{calculators}, including target quantities such as the SIRS score from Figure~\ref{fig:problem}.
In the original study, average scores were reported across all calculators: 37.9\% in the zero-shot setting with GPT-4 and 50.9\% in the one-shot setting. In the latter, the demonstration always used the same calculator as the test case, thereby evaluating the model’s ability to extract data and reproduce an identical reasoning chain.


For our input, we concatenate the patient note and the original question, followed by a concise definition of the relevant formulas or rules.
In the long-context CoT setting (for SSRMs), this concatenation serves as the sole input.
In prompt-based settings, we employ a single two-shot CoT demonstration involving calculations of the same \emph{type} (formula or rules), though not necessarily the same \emph{calculator}, thereby testing the LLM's ability to extract data and perform a potentially different calculation.
Therefore, MedCalcV2 scores are not directly comparable to those of MedCalcBench.


We implement two additional changes.
First, we remove training samples in the original MedCalcBench that overlap with the test data to ensure a clean evaluation.
Second, during testing, we discovered errors in results for the Glasgow Coma Scale Calculator: each ground-truth explanation duplicates the verbal-response rule and erroneously adds its value twice, leading to incorrect final scores.
We manually correct these errors by deleting the duplicate lines and adjusting the final values in both the ground-truth explanations and the expected outputs.
MedCalcV2 will be made available.

\subsection{Typicality Audit Configuration}\label{app:audits}


Results labeled HMM* are obtained via a grid search over hidden-state counts (1, 2, 5, 10) and n-gram sizes (1, 2, 3, 10, 25, 50), selecting the model with the lowest Bayesian Information Criterion (BIC) score~\citep{dridi2018akaike}.
HMM are implemented using the \texttt{CategoricalHMM} class from \texttt{hmmlearn}, with preprocessing to convert sequences into n-gram representations.
Each sequence is augmented with start and end tokens, an unknown-word token, and padded to a uniform length.
We use the Fisher’s exact test in \texttt{scipy.stats} for statistical significance of proportional differences.  

\subsection{Prompt for LLM-Generated Audits}
Generated audits are created by prompting Claude-Sonnet-4-20250514 using the following prompt, replacing the label [TASKNAME] with the name of the task the audits are being generated for.
\begin{figure*}[htbp]
\scriptsize
\centering
\begin{tcolorbox}[colback=white, colframe=customBlue, width=1.0\textwidth, arc=3mm, boxrule=0.5mm, title=Prompt for LLM-Generated Audits]
\begin{Verbatim}[breaklines=true, breakanywhere=true, formatcom=\bfseries]
The attached file 'Example Audits' contains examples of audit functions which run on the traced outputs of functions called mocks. Each audit function tests the output to ensure that the mock has been run correctly by testing individual parts of the traced output, ensuring that each function the mock expects has been called, that the correct outputs lead to the correct inputs, and so on.

The attached file 'audit.py' contains the code which runs audit functions. Use this file to reference the expected structure of the dataframe that audit functions call on.

The attached file 'Audit Targets for [TASKNAME]' contains several traced outputs for a mock function , [TASKNAME]. Generate a set of audit functions matching the format and construction of the examples from 'Example Audits', which will test other traced outputs of the function [TASKNAME]. Your generated audits should not programmatically generate the messages for success or failure.

Return only the python code for your output, with no extraneous introduction or afterward. Do not encase your output in backticks. Make sure to include imports and an if-main function.
\end{Verbatim}

\end{tcolorbox}
\caption{Prompt for LLM-Generated Audits.}
\label{appendix_llm_generated_audits_prompt}
\end{figure*}

\subsection{Additional results}\label{appendix_additional_results}

\input{tables/struct_audit_medcalc_prompted_full}

\input{tables/gen_audit_bbh}
Table~\ref{tab:structaudit_prompt_full} presents the comparison between the prompted Sonnet 3.5 model and a smaller prompted model, \texttt{Qwen2.5-7B-Instruct}, which is similar to the model we trained.
The structured audits reveal that \texttt{Qwen2.5-7B-Instruct}'s performance diverges significantly from the larger Sonnet model.
In the Formula task, Sonnet 3.5 exhibits no significant reasoning errors, whereas \texttt{Qwen2.5-7B-Instruct} frequently commits errors in the initial reasoning steps, resulting in substantially poorer outcomes.
In the Rule task, \texttt{Qwen2.5-7B-Instruct} demonstrates a distinct failure mode than Sonnet model: it generates correctly structured solution traces, but then fails to execute each individual step correctly.

Table~\ref{tab:bbh} shows results with LLM-generated audits on 21 tasks from the BBH benchmark suite. 
We report the number of lines of code in the generated audits, and the average number of audits that are run on each example.
As a concise measure of the utility of the audits, we report the smallest $p$-value of any audit, as computed in Table~\ref{tab:structaudit} (i.e., for the null hypothesis that audit failure is not associated with incorrect outputs.)  A small $p$-value indicates that some LLM-generated audit does indeed provide information about an ``interesting'' reasoning failure.  Nearly half of the generated audits have $p$-values less than 0.05, including all four of the tasks with the highest error rates.

\input{tables/gen_struct_prompted_medcalc}

Table~\ref{tab:gen-struct-prompt-medcalc} shows results of generated structured audits on the same reasoning traces used in Table~\ref{tab:structaudit_prompt}.

\subsection{Token Usage Analysis}\label{appendix_token_usage}
\begin{wraptable}{r}{0.5\textwidth}
\vspace{-1.2em}
\centering
\resizebox{\linewidth}{!}{
\begin{tabular}{lcc}
\toprule
\textbf{Dataset} & \textbf{Qwen Unstructured} & \textbf{Qwen SSRM} \\
\midrule
GSM8K              & 319.78  & 841.72  \\
Math500            & 909.27  & 978.89  \\
MedCalcV2 Formulas & 411.80  & 1778.14 \\
MedCalcV2 Rules    & 425.70  & 2260.87 \\
GPQA Diamond       & 1608.33 & 1411.29 \\
MedQA              & 359.25  & 1065.34 \\
\bottomrule
\end{tabular}
}
\caption{Token usage of Qwen SSRM and corresponding unstructured baseline across datasets.}
\label{tab:token_usage}
\end{wraptable}
As shown in Table~\ref{tab:token_usage}, SSRM consumes more tokens than the unstructured reasoning baselines on MedCalcV2 Rules and Formulas, whereas token usage is comparable on MATH500 and GPQA-Diamond. The higher token consumption primarily results from redundant arguments and variable referencing, as illustrated in Figure~\ref{fig:approach}. Developing a more efficient variable referencing mechanism is left for future work.

\subsection{Evaluation Configurations}
We use \href{https://github.com/huggingface/lighteval}{\texttt{Lighteval}} for all evaluations.
For non-reasoning models, we report accuracy using greedy decoding.
For reasoning models, we set the temperature to 0.6 and top-$p$ to 0.95.
For the AIME24 dataset---where we observe high variance---we sample 32 responses using a temperature of 0.7 for non-reasoning models, while retaining the configurations for reasoning models, and report Pass@1.

\subsection{Test-Time-Scaling with Typicality Audits}\label{appendix-test-time-scaling}
\input{tables/test-time-scaling}
To investigate the effectiveness of combining test-time-scaling with audits, we apply typicality audits (HMM*).
We perform a grid search using the first half of the generated responses from the benchmark; to ensure data integrity, we evaluate the model only on the second half.
We consider two variants here: vanilla self-consistency and audit-based self-consistency. 
Given a sampling budget of $k$ responses per question, in vanilla self-consistency we sample $k$ times per question and use majority voting to determine the final answer.
In audit-based self-consistency, we divide the model-generated traces into tertiles: for traces in top tertile we perform no additional sampling, for those in the middle tertile we sample $k-3$ additional times, and for those in the bottom tertile we sample $k-1$ additional times.
We report accuracy on the MedCalcV2 Rule tasks, along with the effective number of samples---i.e., the actual number generated under the audit-based procedure. 
For vanilla self-consistency, the total number of samples is $k \times n$, where $n$ is the number of questions in the corresponding benchmark.

As shown in Table~\ref{tab:consistency_comparison}, audit-based self-consistency consistently outperforms vanilla self-consistency given the same per-question sampling budget. 
More specifically, when $k=5$, audit-based self-consistency outperforms vanilla self-consistency by 1.5 percentage points while using only 54.95\% of the total sampling budget.
These preliminary experiments demonstrate the effectiveness of combining typicality audits with test-time-scaling methods and suggest a promising direction for future research.

%% file: tables/hyperparameters.tex
\begin{table}[htbp]
    \footnotesize
    \centering
    \caption{Hyperparameter settings for supervised fine-tuning (SFT) and reinforcement learning with verifiable rewards (RLVR). Both the semi-structured reasoning and CoT baseline settings use the same set of hyperparameters. $\dagger$: max sequence length for SFT and max generation length for RLVR.}
    \label{tab:hyperparameter}
    \setlength\tabcolsep{10pt}
    \renewcommand{\arraystretch}{1.2}
    \begin{tabular}{@{}lcc@{}}
        \toprule
        \textbf{Hyperparameter} & \textbf{SFT} & \textbf{RLVR} \\ 
        \midrule
        Optimizer                   & AdamW        & AdamW          \\
        Actor Learning Rate         & 1e-5         & 1e-6           \\
        Weight Decay                & 1e-4         & 0.1            \\
        Warmup Ratio                & 0.1          & 0.01           \\
        Prompt Length               & -            & 2048           \\
        Max Length\textsuperscript{$\dagger$}                  & 16384        & 4096           \\
        Loss Agg Mode               & -            & token\_mean    \\
        Grad Clip                   & 0.2          & 1.0            \\
        Batch Size                  & 128          & 256            \\
        MiniBatch Size              & -            & 256 (On-Policy) \\
        Num Responses Per Prompt    & -            & 8              \\
        Temperature                 & -            & 1.0            \\
        Sequence Packing            & False        & True           \\
        Entropy Coeff               & -            & 0.0            \\
        KL Loss Coeff               & -            & 0.0            \\
        Epochs                      & 5            & 10             \\
        \bottomrule
    \end{tabular}\label{tab:appendix_hyperparameters}
\end{table}

%% file: tables/struct_audit_medcalc_prompted_full.tex
\begin{table}[tb]

\caption{Results of applying hand-coded structured audits to prompted models for MedCalcV2 tasks.}\label{tab:structaudit_prompt_full}

\centering
\resizebox{\linewidth}{!}{
\begin{tabular}{lrrrrrl}
\toprule
 & & \multicolumn{3}{c}{$-$ accuracy and difference $-$} \\
&  \%Failed & Failing & Passing & $\Delta$ & $p$-val & description of audit \\
\midrule
\underline{MedCalcV2 Formulas}
& 1.712 & 0.000 & 0.662 & 0.662 & 0.162 & one "get\_data" step \\
& 2.055 & 0.000 & 0.664 & 0.664 & 0.086 & one "insert\_variables" step \\
~~Claude 3.5
& 3.767 & 0.091 & 0.673 & 0.582 & 0.033 & "solve\_formula" output is a number \\
~~(65.1\% acc)
 & 9.247 & 0.593 & 0.657 & 0.064 & 0.870 & "solve\_formula" output is formatted correctly \\
 & 47.260 & 0.667 & 0.636 & -0.030 & 0.852 & "solve\_formula" math is correct \\
\cline{2-7}\addlinespace[1pt]
& 3.425 & 0.000 & 0.296 & 0.296 & 0.126 & "solve\_formula" output is a string \\
 & 5.479 & 0.188 & 0.292 & 0.104 & 0.777 & "solve\_formula" output is a number \\
 & 7.192 & 0.381 & 0.279 & -0.102 & 0.487 & "solve\_formula" output is formatted correctly \\
~~Qwen2.5-7B-Instruct
 & 29.110 & 0.376 & 0.249 & -0.128 & 0.140 & "solve\_formula" math is correct \\
 ~~(28.6\% acc)
 & 29.795 & 0.103 & 0.365 & 0.261 & 0.000 & one "get\_data" step \\
 & 30.137 & 0.102 & 0.366 & 0.264 & 0.000 & one "insert\_variables" step \\
 & 33.219 & 0.165 & 0.347 & 0.182 & 0.015 & one "analyze\_input" step \\
\midrule
\underline{MedCalcV2 Rules} 
& 5.789 & 0.182 & 0.399 & 0.218 & 0.181 & analyze\_input returns two values \\
 & 14.737 & 0.196 & 0.420 & 0.223 & 0.028 & one step per rule with step\_fn of "convert\_units" \\
~~Claude 3.5 
 & 14.737 & 0.196 & 0.420 & 0.223 & 0.028 & one step per rule with step\_fn of "get\_data" \\
~~(38.7\% acc)
 & 15.789 & 0.183 & 0.425 & 0.242 & 0.015 & one step per rule with step\_fn of "check\_rule" \\
 & 17.105 & 0.169 & 0.432 & 0.263 & 0.005 & one step per rule with step\_fn of "accumulate\_score" \\
\cline{2-7}\addlinespace[1pt]
 & 1.316 & 0.400 & 0.309 & -0.091 & 0.672 & one step per rule with step\_fn of "get\_data" \\
 & 1.579 & 0.333 & 0.310 & -0.023 & 1.000 & one step per rule with step\_fn of "convert\_units" \\
~~Qwen2.5-7B-Instruct 
& 1.579 & 0.333 & 0.310 & -0.023 & 1.000 & one step per rule with step\_fn of "accumulate\_score" \\
~~(31.1\% acc)  
 & 1.579 & 0.333 & 0.310 & -0.023 & 1.000 & one step per rule with step\_fn of "check\_rule" \\
 & 2.632 & 0.500 & 0.305 & -0.195 & 0.363 & one step with step\_fn of "analyze\_input" \\
& 4.737 & 0.389 & 0.307 & -0.082 & 0.630 & analyze\_input returns two values \\
\bottomrule

\end{tabular}
}

\end{table}

%% file: tables/gen_audit_bbh.tex
\begin{table}[!htb]
  \centering
  \caption{Summary of LLM-generated audits on BBH tasks, using a prompted Claude Sonnet 3.5.}
  \label{tab:bbh}

\begin{tabular}{lrrrc}
\toprule
Task	& Task Acc	& Avg Audits/Example	& Code Lines	& Min $p$-value \\
\midrule
geometric shapes	& 37.89\%	& 14.50	& 128	& $<0.001$ \\
formal fallacies	& 46.31\%	& 10.75	& 107	& $<0.001$ \\
causal judgement	& 57.48\%	& 11.75	& 88	& $<0.001$ \\
dyck languages	& 64.00\%	& 27.00	& 89	& $<0.001$ \\
disambiguation qa	& 82.63\%	& 10.98	& 93	&  \\
ruin names	& 83.16\%	& 10.00	& 105	&  \\
penguins in a table	& 87.21\%	& 11.01	& 114	& $<0.05$  \\
multistep arithmetic two	& 87.89\%	& 12.00	& 95	&  \\
snarks	& 91.53\%	& 85.72	& 109	&  \\
date understanding	& 87.89\%	& 11.33	& 88	& \\
logical deduction three objects	& 87.89\%	& 10.99	& 94	& \\
movie recommendation	& 91.05\%	& 13.98	& 90	&  \\
reasoning about colored objects	& 94.21\%	& 14.00	& 95	& \\
word sorting	& 95.26\%	& 19.82	& 110	& $<0.05$ \\
boolean expressions	& 95.26\%	& 6.09	& 92	&$<0.05$ \\
temporal sequences	& 96.84\%	& 12.98	& 90	& \\
sports understanding	& 97.37\%	& 7.00	& 71	& $<0.05$ \\
hyperbaton	& 97.89\%	& 7.00	& 69	& $<0.001$ \\
tracking shuffled objects	& 98.95\%	& 17.00	& 105	& $<0.05$ \\
object counting	& 100.00\%	& 9.00	& 70	&  \\
web of lies	& 100.00\%	& 15.00	& 94	& \\
\bottomrule
\end{tabular}

\end{table}

%% file: tables/gen_struct_prompted_medcalc.tex
\begin{table}[htbp]

 \vspace{1em}

\caption{LLM-generated structured audits Claude Sonnet 3.5 Prompted Models for MedCalcV2.} \label{tab:gen-struct-prompt-medcalc}

 \vspace{-0.5em}

\begin{center}
\resizebox{\linewidth}{!}{
\begin{tabular}{lrrrrrrl}
\toprule
 & & \multicolumn{3}{c}{$-$ accuracy and difference $-$} &  \\
&  \%Failed & Failing & Passing & $\Delta$ & $p$-val & description of audit \\
\midrule
Formulas	& 1.71	& 0.00\%	& 44.60\%	& 44.60\%	& *	& one get\_data step \\
	& 0.34	& 0.00\%	& 43.99\%	& 43.99\%	& 		& one analyze\_input step \\
	& 0.68	& 0.00\%	& 44.14\%	& 44.14\%	& 		& analyze\_input returns tuple with 2 elements \\
	& 2.05	& 0.00\%	& 44.76\%	& 44.76\%	& *		& one insert\_variables step \\
	& 4.79	& 35.71\%	& 44.24\%	& 8.53\%	& 		& convert\_units called on each datapoint \\
	& 3.42	& 50.00\%	& 43.62\%	& -6.38\%	& 		& convert\_units' second input is a datapoint \\
	& 0.68	& 50.00\%	& 43.79\%	& -6.21\%	& 		& convert\_units's first input is the formula \\
	& 0.68	& 50.00\%	& 43.79\%	& -6.21\%	& 		& insert\_variables' first input is the formula \\
	& 3.08	& 44.44\%	& 43.82\%	& -0.63\%	& 		& insert\_variables' second input is an output of convert\_units \\
	& 2.74	& 37.50\%	& 44.01\%	& 6.51\%	& 		& get\_data's inputs match the output of analyze\_input \\
	& 0.34	& 0.00\%	& 43.99\%	& 43.99\%	& 		& solve\_formula's input is an output of insert\_variables \\
	& 92.81	& 42.80\%	& 57.14\%	& 14.34\%	& 	 	& final answer matches last solve\_formula output \\
\midrule
Rules	& 5.79	& 18.18\%	& 39.94\%	& 21.76\%	& *	 	& analyze\_input returns tuple with 2 elements \\
	& 14.74	& 19.64\%	& 41.98\%	& 22.33\%	& **		& get\_data called for each rule \\
	& 14.74	& 19.64\%	& 41.98\%	& 22.33\%	& **		& consistent rules across get\_data steps \\
	& 1.05	& 0.00\%	& 39.10\%	& 39.10\%	& 	 	& convert\_units inputs are outputs of get\_data \\
	& 14.74	& 19.64\%	& 41.98\%	& 22.33\%	& **	 	& convert\_units called for each rule \\
	& 14.74	& 19.64\%	& 41.98\%	& 22.33\%	& **	 	& consistent rules across convert\_units steps \\
	& 2.63	& 0.00\%	& 39.73\%	& 39.73\%	& **		& check\_rule inputs are outputs of convert\_units \\
	& 15.79	& 18.33\%	& 42.50\%	& 24.17\%	& **	& check\_rule called for each rule \\
	& 15.79	& 18.33\%	& 42.50\%	& 24.17\%	& **	 	& consistent rules across check\_rule steps \\
	& 0.79	& 0.00\%	& 38.99\%	& 38.99\%	& 	& accumulate\_score inputs are outputs of check\_rule \\
	& 17.11	& 16.92\%	& 43.17\%	& 26.25\%	& **		& accumulate\_score called for each rule \\

\bottomrule

\end{tabular}
}

\end{center}
\vspace{-1.5em}
\end{table}

%% file: tables/test-time-scaling.tex
\begin{table}[htbp]
    \footnotesize
    \centering
    \caption{Comparison of Self-Consistency and Audit-Based Self-Consistency on MedcalcV2 Rule.}
    \label{tab:consistency_comparison}
    \begin{tabular}{@{}lccc@{}}
        \toprule
        \textbf{Sampling Budget} & \textbf{Self-Consistency} & \textbf{Audit-Based Self-Consistency} & \textbf{Effective Samples} \\ 
        \midrule
        Greedy (Temp = 0)          & 44.2                      & 44.2                            & -                          \\
        Sampling (Temp = 0.7)      & 44.2                      & 44.2                            & -                          \\
        3       & 46.3                      & 45.3                            & 306 (53.68\%)              \\
        5       & 45.3                      & 46.8                            & 522 (54.95\%)              \\
        7       & 45.3                      & 45.3                            & 764 (57.44\%)              \\
        9       & 45.3                      & 45.3                            & 1002 (58.60\%)              \\
        15      & 45.3                      & 44.2                            & 1702 (59.72\%)             \\
        30      & 45.2                      & 46.3                            & 3501 (61.42\%)             \\
        60      & 44.7                      & 45.8                            & 7071 (62.03\%)             \\
        \bottomrule
    \end{tabular}
\end{table}